\documentclass{llncs}

\usepackage[pdftex,dvipsnames,table]{xcolor}  % Coloured text etc.
\usepackage{xargs} % Use more than one optional parameter in a new commands
\usepackage[colorinlistoftodos,prependcaption,textsize=small]{todonotes}
\newcommandx{\unsure}[2][1=]{\todo[linecolor=red,backgroundcolor=red!25,bordercolor=red,#1]{#2}}
\newcommandx{\change}[2][1=]{\todo[linecolor=blue,backgroundcolor=blue!25,bordercolor=blue,#1]{#2}}
\newcommandx{\info}[2][1=]{\todo[linecolor=OliveGreen,backgroundcolor=OliveGreen!25,bordercolor=OliveGreen,#1]{#2}}
\newcommandx{\improvement}[2][1=]{\todo[linecolor=Plum,backgroundcolor=Plum!25,bordercolor=Plum,#1]{#2}}
\newcommandx{\thiswillnotshow}[2][1=]{\todo[disable,#1]{#2}}
\newcommandx{\thl}[2][1=]{\todo[linecolor=blue,backgroundcolor=yellow!25,bordercolor=yellow,#1]{#2}}
\usepackage{soul}
\renewcommand\hl[1]{#1} % <<<<<<<<<<<<<<<<<<<<<<<<<<<<<<<<<<<<<

\begin{document}

\frontmatter          % for the preliminaries

\pagestyle{headings}  % switches on printing of running heads
%\addtocmark{Hamiltonian Mechanics} % additional mark in the TOC

\mainmatter              % start of the contributions

\title{Towards Long-Term Memory for Social Robots: Proposing a New Challenge for the \hl{RoboCup@Home League}}
%
%\titlerunning{Hamiltonian Mechanics}  % abbreviated title (for running head)
\author{Mat\'ias Pavez\inst{1} \and Javier Ruiz del Solar\inst{1} \and
Victoria Amo\inst{2} \and Felix Meyer zu Driehausen\inst{2}}
\authorrunning{Mat\'ias Pavez et al.} % abbreviated author list (for running head)
\tocauthor{Mat\'ias Pavez, Javier Ruiz del Solar, Victoria Amo, and Felix Meyer zu Driehausen}
\institute{
	Advanced Mining Technology Center \& Dept. of E.E., Universidad de Chile
\and
	Institute of Cognitive Science, Universit\"{a}t Osnabr\"{u}ck\\
	\{\email{matias.pavez@ing.uchile.cl}, \email{jruizd@ing.uchile.cl}\\
	\email{vamo@uos.de}, \email{fmeyerzudrie@uos.de}\}
}

\maketitle              % typeset the title of the contribution

\begin{abstract}
% at least 70 and at most 150 words. 
\hl{Long-term memory is essential to feel like a continuous being, and to be able to interact/communicate coherently. Social robots need long-term memories in order to establish long-term relationships with humans and other robots, and do not act just for the moment.}  In this paper this challenge is highlighted, open questions are identified, the need of addressing this challenge in the RoboCup@Home League with new tests is motivated, and a new test is proposed.
\keywords{\hl{long-term-memory, service robot, social robot, RoboCup@Home}}
\end{abstract}

\section{Introduction}

Long-term memory allows humans to feel continuous and coherent in his/her thoughts, i.e., to be a continuous person with a continuous life. Hence, long-term memory is an essential component of social interaction between people in daily life, it allows remembering names, events, duties, relationships, etc. In fact, when long-term memory does not work properly due to illness (e.g., Alzheimer disease), the ability to interact with other human beings is severely damaged.

Therefore, it seems evident that a key aspect in achieving long-term interaction and social relationship between humans and social robots is the requirement of a long-term memory system for the latter. Memory constitutes an important part of a
cognitive system implementation; in fact, it is the link of prior experiences with
ongoing and future behavior. As stated by \cite{1_wood}, a nontrivial level of social interaction requires that the robot should be able to use both semantic and episodic information. Both, semantic and episodic memories constitute the declarative long-term memory. While semantic memory is a repository for facts, such as knowing that the capital of Chile is Santiago, episodic memory is the memory of past experiences, i.e., remembering events and their associated context (places, persons, emotions, etc.). Both kinds of memories work together, and both need to be implemented for providing appropriate social skills for social robots. However, in contrast to implementing semantic memories, the implementation of episodic memories for social robots is much less developed, and therefore we focus our work on the latter.

The amount of information to be processed in a lifetime is vast; therefore, efficient methods are required for acquiring, filtering, encoding, storing, deleting and updating a robot's episodic knowledge of its working environment. This information can be encoded in symbolic form and held in a storage module invoking the functionality of an episodic memory system. So far, this challenge has not been addressed appropriately, and the construction of long-term memories for social robots, which include both episodic and semantic components, has not been achieved. In this context, the main goals of this paper are: (i) to highlight the need of building appropriate long-term memory for social robots, (ii) to identify open questions that need to be answered in order to fulfill this first goal, and (iii) to provide basic concepts to address this challenge in the RoboCup@Home League with new tests. We propose the structure of such kind of tests, and give an example of a concrete test. This paper is organized as follows: The basic aspects of human memory are summarized in Section 2. Relevant open questions are identified in Section 3. The structure of the proposed tests for the RoboCup@Home League is presented in Section 4. Finally, some conclusions of this work are outlined in Section 5.

\section{Human Memory}

According to \cite{6_Eichenbaum:2008}, memory is ``the record of experience presented in the brain''. There are multiple memory systems that work in a complementary way. These systems have different  functions, and they are characterized by different operating characteristics and brain structures in which they are embodied.

A first categorization is related to the persistency of the stored information. The \textit{Sensory Memory} is able to store information acquired by our sensory systems just for the fraction of a second. This information then enters the so-called \textit{Short-Term Memory}, which support ``brief storage and immediate recall of substantial detail'' \cite{6_Eichenbaum:2008}. Part of the stored information is then consolidated into the \textit{Long-Term Memory}. The term \textit{Working Memory} is sometimes used instead of Short-Term Memory, although the working memory concept includes processes and structures used for the temporal storage and manipulation of information \cite{5_Vijayakumar2014}.

The Long-Term Memory is composed by a \textit{Declarative or Explicit Memory}, which refers to information that is remembered consciously, and a \textit{non-Declarative or Implicit Memory}, which refers to skills, abilities and tasks that can remembered implicitly (e.g., how to ride a bicycle). The Implicit Memory is further divided in \textit{Procedural Memory}, in charge of storing procedures or ways of doing tasks, and \textit{Priming}, which refers to the fact that some experiences are primed or recalled when a given stimuli is received. The Declarative Memory is divided into two complementary memories, the \textit{Episodic Memory} and the \textit{Semantic Memory}. Episodic memory, first defined by Tulving \cite{2_tulving}, refers to the memory of specific events occurring at a specific place and time and enables human beings to remember past experiences. Semantic memory is basically a repository for facts. Emotional experiences are also stored in the brain and they have influence on how other information is stored. The \textit{memory about emotions} is stored in the declarative memory, while the \textit{emotional memory} belongs to the implicit memory. Detailed explanations about the episodic, semantic and emotional memories can be found in \cite{8_2008iii,7_eichenbaum2008learning}.

The main functionalities provided by the declarative long-term memory are: (i) the capability of remembering facts, concepts, events, experiences, skills, tasks, emotions, (ii) the ability to feel like a continuous person, (iii) a way of linking prior experiences with ongoing and future behavior, and (iv) the ability to interact with other human beings, i.e. be able to communicate coherently, and to build long-term relationships.

In this article our analysis will focus in the episodic memory, which as already mentioned, is related to the conscious remembrance of context-dependent events that are personally experienced. By context-dependent is it meant the cognitive state \textemdash the temporal, spatial and emotional/affective context \textemdash, as well as the embodied nature of the experience, i.e. the sensory-perceptual processing of a given experience \cite{9_philip}. For instance, the experience/event of visiting your mother's house for having dinner last night includes a temporal context (last night), a spatial context (your mother's house), an emotional context (spending time nicely with your mother), and sensory-perceptual experiences (how the dinner tasted).

\section{Open Questions for Implementing Long-Term Memory for Social Robots}

During the past decade there have been several approaches that implemented episodic and semantic long-term memory in artificial systems \hl{(e.g., }\cite{16_Deutsch2008,13_Dodd2005,15_Jockel2008,14_Kuppuswamy2006ACC,11_LAIRD19871,Leconte:2016,10_Nuxoll2004ACM,12_1513818,5_Vijayakumar2014}\hl{)}. We believe that these works addressed only partially some of the main challenges that poses the implementation of long-term memory for social robots, and that still some of the questions described in the next paragraphs need to be answered.

\textit{How to store events and experiences in the form of episodes in the episodic memory?} It is still not clear how the cognitive state \textemdash temporal, spatial and emotional/affective context \textemdash associated to events, as well as the embodied nature of them, can be stored by a social robot. Naturally, the temporal context can be easily stored (e.g., using time stamps). The ability to determine and store spatial context has advanced largely in the last few years thanks to the deep learning revolution that allows the recognition of places, objects, and persons more easily. Yet, this has so far not been implemented in social robots that interact continuously with the changing world. Very few works have addressed the task of storing the emotional context and the embodied nature of the experience.

\textit{How to give different levels of relevance to the different episodes in terms of its novelty or the associate emotional state?} As in the case of human beings, the stored episodes have different levels of relevance, which depends in the associate emotional state, among many other factors. Mechanisms for determining autonomously this level of relevance need to be developed.

\textit{How to store emotional states?} In the human brain emotional situations are stored in the explicit and implicit memory systems, and experiences with a strong emotive content produce powerful and vivid memories. Mechanisms for implementing these functionalities need to be developed.

\textit{How to consolidate short-term memory into the long-term memory?} The update of the long-term memory is a complex and time-consuming process that involves the consolidation of short-term memories into long-term ones \cite{3_bailey}. In the case of humans this is carried out during the sleep process, and consumes a large amount of brain resources \cite{4_walker}. Therefore, the update of episodic information must be carefully designed and implemented in the robot case. No work has addressed the challenge of long-term memory consolidation for a social robot operating continuously in the real-word.

\textit{Which mechanisms to use for forgetting and repression?} Given that the memory capacity is limited, forgetting mechanisms need to be implemented on basis of the relevance of the stored information. In addition, it must be analyzed if, as in the case of human beings, it is required to implement repression mechanisms that unconsciously block memories in order to protect the self from situations/emotions that she/he cannot cope with \cite{18_Ho2009}.

\textit{How to address the ethical issues related to the management of personal information of human beings acquired by social robots?} A social robot with long-term memory will store information related with his/her human mates. It must be analyzed how this personal information will be protected, managed, and eventually, deleted.

\section{Long-Term Memory in the \hl{RoboCup@Home League}}

In this Section we propose a new test for the RoboCup@Home league. First, we will describe the minimum requirements for an EpLTM (Episodic Long-Term Memory) implementation, how to validate them, and which sources of information are valid when generating memories. \hl{Finally, we present a test proposal for the competition}, focused on EpLTM, Human-Robot Interaction (HRI) and perception.

\subsection{Requirements}

At present, there is no consensus on the way an EpLTM should be implemented for service robots, but many approaches can be found  \cite{18_Ho2009,15_Jockel2008,Kelley2014,KimMinJoo2016,17_spexard,21_Stachowicz2012,5_Vijayakumar2014}. As the RoboCup’s goal is to boost research, we propose to evaluate only the core requirements an EpLTM must fulfill, avoiding to force an specific implementation on the teams. The proposed requirements can be separated into 2 categories: I. exclusively episodic, and II. requirements related to historical and emotional relevances.

Category I is build from the \hl{11 design requirements} $\{R_1,\ldots, R_{11}\}$ presented by Stachowicz \cite{21_Stachowicz2012}. These were created to match the characteristics every EpLTM system must satisfy and are the minimum points for validation. \hl{The requirements (R1, R2, R4) declare that every episode must be recollected and stored by its spatio-temporal context: \textit{what}, \textit{when}, and \textit{where} it happened. Moreover, there are no restrictions on which information the \textit{what} field can contain; For the competition there are useful pieces of data to remember, for example, static information about known people or objects (name, age), and their dynamic state (last location, clothes, emotions). On the other hand, (R3, R6) state how the \textit{what} field can be accessed and modified, while (R5, R7, R8) give some rules about the episode system structure (children episodes, anidation and transposition).} In this work, R9 (non intrusiveness), R10 (efficiency), and R11 (scalability) are left out, because they relate to desirable design requirements and are not considered as candidates for validation during a test.

Category II adds the concept of relevance to each episode in memory, which is not covered by category I. Episodic relevance is essential when remembering interesting events, allowing access to episodes by their importance. On the one hand, we propose the historic relevance, which is directly related to the age of an episode; the lower its antiquity is, the higher its importance, which means a high probability to remember recent events. On the other hand, we propose the emotional relevance by assigning an emotion and its related magnitude to each episode; this allows to retrieve older but important events. \hl{It is important to stress that knowing \textit{why} the episode is relevant and \textit{how} the emotion relates to the episode is not required, as this depends on the emotion engine implementation.}

\subsection{\hl{Required information for validation}}

Next we present a proposal for \hl{the minimum data required to be stored for the competition, and the level of detail needed when validating the requirements (categories I and II)}. The concepts to be delimited are the episode definition, its contents (What, Where, When), and the associated emotions. 

\hl{Although Stachowicz's requirements do not impose the exact data to be gathered for (What, Where, When) fields, a set of verifiable entities should be defined for the competition. This serves as a way to normalize the validation process, by clarifying which data and format will be required for validation. It is important to highlight that by the following constraints we expect not to impose an implementation to the teams, but just to formalize the minimum required capabilities to any robot in competition.}

\subsubsection{Episodes:} In order to provide context to any episodic query or to enable a precise memory description by the robot, \hl{we propose at least the following nested episodic levels:}
\begin{enumerate}
\item \textit{Context:} RoboCup, Stage X, Test Y, Subtest Z. This let us identify generally the spatio-temporal context of an episode. Subtest Z only applies to tests where sub stages are clearly defined, as in: ``Stage 1, Test: SPR, Subtest: The Riddle Game''.
\item \textit{Tasks:} There must exist an episode related to each task or order executed by the robot. \hl{Tasks are defined inside a \textit{Context} or inside other \textit{Task}.}
\item \textit{Capabilities:} There must exist an episode related to each high-level robot capability: navigation, manipulation, perception, HRI. \hl{\textit{Capabilities} are defined inside a \textit{Task}.}
\end{enumerate}

\subsubsection{When:} \hl{Just knowing the sequence of episodes is not enough for validation. On the one hand, transposed episodes, i.e. episodes that are simultaneous, cannot be sequenced; on the other hand, the referees need a way to verify that the given episode description relates to the recorded time. Consequently, at least, the temporal information of an episode must consider initial and final timestamps.} These can be described in terms of minutes, hours, days, weeks, months or years. 

\subsubsection{Where:} Location must be described in a simple way. Using coordinates like (x,y,z) is not allowed. What is allowed:
\begin{itemize}
\item The robot can show a map of the arena with drawings marking the interesting locations.
\item The robot can describe locations using semantic information, with room names and elements of the arena. Some examples: Inside/Outside the arena, rooms (kitchen, bedroom), furniture (desk, fridge), or by using relative positions (at the left of, over the). For tests which require Simultaneous Localization and Mapping (SLAM) the description can be in terms of known areas, as the ``bar'' in the Restaurant test.
\end{itemize}

\subsubsection{What:} \hl{As the information to store is not clearly defined, this highly depends on the team implementation. However, at least some entities and fields should be defined by the referees/technical committee, in order to give a normalized base for all queries and validations during a test. Proposed entities to be stored are: people, objects and locations. Fields for each of these can be obtained from capabilities required on previous tests (e.g., age/emotion recognition, last seen location, and face images).} Event description can be made verbally or by displaying a graph with the related sub-episodes.
\begin{itemize}
\item The verbal description is preferred. This should be related as a story, by saying the associated episode sequence.
\item The description should be as specific as possible. E.g.: ``I moved'' vs. ``I moved towards the door''.
\item Transposed episodes should only be considered if they are in context. \hl{These might require more complex verbal explanations to emphasize the concurrence of the actions.}
\end{itemize}

\subsubsection{Emotions:} \hl{How emotions are generated for a given robot and how they are associated to any given episode strongly depends on the emotional system used by the team}.\hl{Therefore,} no matter what emotional model is used, we propose to restrict the emotions to just 4 groups: Joy/Trust, Sadness/Fear, Surprise/Anticipation, Anger/Disgust. These are obtained from \hl{Plutchik's} theory of emotions \cite{23_plutchik}. This selection is made to have a simple and verifiable set of emotions for the competition. Each episode must be related to at least one emotion and its intensity. Emotion intensity must have at least a resolution of 4 levels: ``normal'', ``a little \textless happy\textgreater'', ``\textless happy\textgreater'', ``very \textless happy\textgreater''.

\subsubsection{Other limitations:} There are no proposed rules on how the memories are stored into semantic memory, as it depends on the implementation. There are no rules associated to other concepts.

\subsection{Episode Generation}

There are many sources from where the robot can generate episodes for the EpLTM: preparation for the competition, the travel, setup days, time between tests, \hl{non-RoboCup} related episodes and the tests in which the robot participated. However, only the last one is a verifiable source of information. Then, we propose to limit the episodes to only the ones related to the tests of the current competition. On the first hand, this lets us simplify the evaluation and veracity of the memories. On the second hand, by only considering these episodes, new participants will compete in the same terms as older teams, this also serves as a regulation between the amount of memories gathered by teams which compete many times in a year.

\hl{As the competition lasts only a few days, the number of learning instances to gather episodes is small.} In order to increase the amount of episodes, we propose two approaches. First and most important, the EpLTM test must be postponed as much as possible, ideally as the last test of Stage 2. The second proposal is to add a ``Memory Setup'' stage at the end of the ``Setup Days'' period, where each robot can generate interesting memories for the queries of the test. E.g.: By maintaining an informal conversation with someone of the committee or by an introduction to the people that it will find through the tests.

Depending on the queries the robot will encounter on the competition, the organizing committee can find new requirements. For instance for queries about people participating in 2 or more tests, the same name should be assigned to them, and they should wear similar but not identical clothes.

Finally, we have proposed to only consider episodes related to the current RoboCup competition, but it is important to mention that adding other events and older RoboCup competitions has some advantages. This directly affects the amount of episodes the robot will recollect. \hl{Moreover, this enables us to consider queries based on episodic inference, so the robot can be confronted to tasks requiring extrapolation from similar situations (e.g., Joe usually wants me to clean up the table after a meal). This capability can be added in later competitions, when the EpLTM test is considered solved.}

\subsection{Validation}

The introduction of the memory concept is susceptible to cheating, for example, with manually written episodes during the competition or through a random episode generator. To attack this  problem, first we consider the Fair Play concept. On the other hand, a strategy to hinder hardcoding is to increase the number of episodes and increase the number of available queries. However, the simple solution is to require evidence of each described episode.

\begin{figure}
	\centering
	\includegraphics[height=6.2cm]{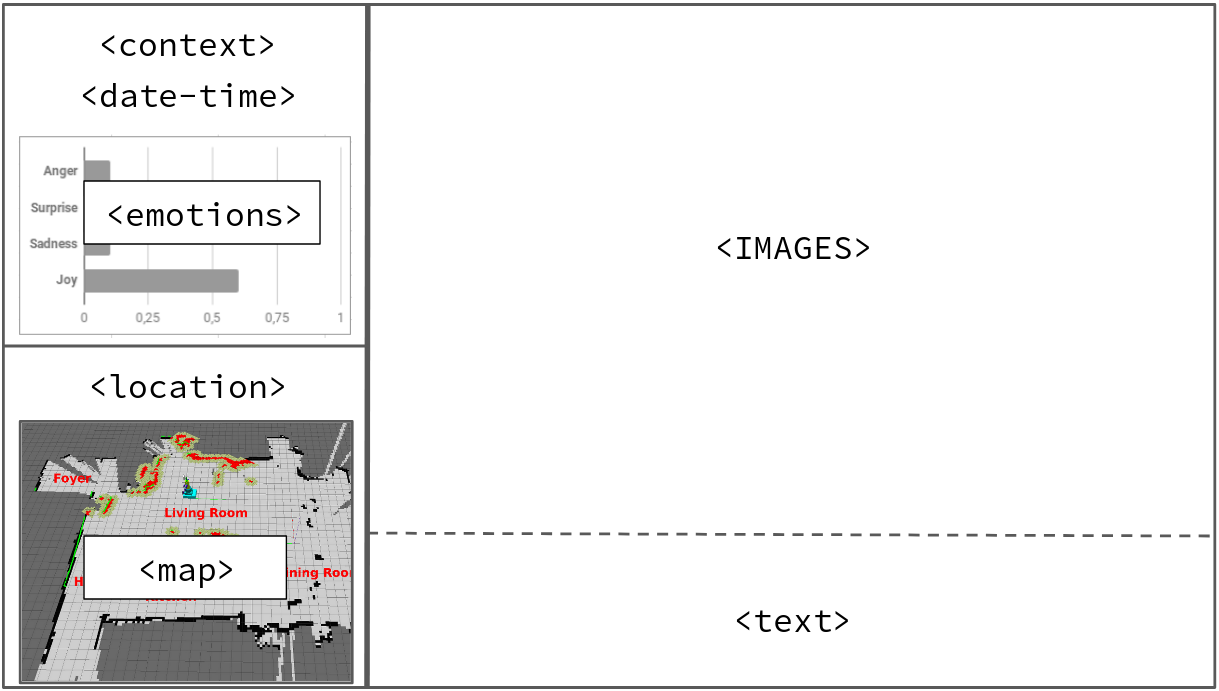}
	\caption{\hl{Example of episodic visualization for the competition with fields showing all verifiable information by the referees. In the upper-left panel, the \textit{context} and \textit{date-time} are shown (e.g., ``EpLTM Test'' - ``Wed, July 17, 2019  15:40:15''), followed by a graph with the emotion intensities. In the lower-left panel, location information is displayed using a \textit{map} of the environment and the \textit{location} name (e.g., ``kitchen''). In the right panel, the \textit{images} and \textit{text} fields are meant to display images recorded by the robot, interaction subtitles, and other useful information as proof for validation.}}
	\label{fig:vizbox}
\end{figure}

Evidences fulfill two purposes, they hinder the cheating and also simplify the score assignation by the referees. When describing an episode it is desirable that the robot displays related evidence. As an example, the robot can provide a visualization as the one shown on \hl{Figure }\ref{fig:vizbox}., where location, time, and context are given. \hl{Moreover, this can be displayed through the Vizbox} \cite{22_robocup:rulebook_2017} \hl{application, so that referees and audience can see it.}
\begin{itemize}
\item Task is validated by displaying the requirement as text.
\item Location is validated by displaying a colored map and images.
\item Sub-Episodes are validated with images or video.
\end{itemize}

\subsection{\hl{Test Proposal: }Sick and elderly care}

Next we present an example of an @Home like test with EpLTM requirements. \hl{The test is based on the methodology proposed in this Section. Particularly, it is important to postpone the test as part of Stage 2, so that the robots can collect as much interesting episodes as possible beforehand.} We expect that this proposal can be adjusted as needed for the competition. 

\subsubsection{\hl{Focus:}} The robot must help a sick or elderly person with reduced mobility (in bed/wheelchair), by answering questions about the home and recent events at which he cannot attend. The test is focused on EpLTM, perception and HRI. 

\subsubsection{\hl{Setup:}} The test takes place on the @Home arena. The operator is waiting in the bedroom, lying on the bed or sitting on a chair. The arena keeps the same structure and items as in other tests, but with some small changes. Other people are located in the house with which the robot can interact as needed.

\subsubsection{\hl{Task:}} The robot starts by entering the arena, it moves to the bedroom, approaches the operator and asks if any assistance is needed. The operator explains he is sick/tired and cannot move, so it will make some questions to the robot. After 4 queries, the operator tells the robot to leave the bedroom.

\subsubsection{\hl{Considerations:}}
Queries can be separated into 3 categories, depending on their requirements. The robot should answer at least one question of each category.
\begin{itemize}
\item Cat 1: Queries about memories and emotions.
\item Cat 2: Queries which require investigating objects in the arena.
\item Cat 3: Queries which require interacting with people in the arena.
\item We recommend the use of an episodic queries generator, built to match requirements of categories I and II.
\item The robot must show evidences when answering, e.g., in a screen or using Vizbox \cite{22_robocup:rulebook_2017}, as shown on the official Rulebook.
\item Referees must validate the coherence between answers and provided evidence.
\item Rooms, people and objects must be set up according to the possible queries.
\end{itemize}

\section{Conclusions}

The development of EpLTM for social robots is an important challenge for improving
human-robot interaction, and @Home has the means to boost the progress. For that
reason, we have presented a test proposal focused on EpLTM, its requirements and
validation strategies. The proposal is focused on the minimum requirements any
social robot implementing EpLTM should fulfill, but trying not to impose an specific
implementation to the teams.

\bibliographystyle{splncs03}
\bibliography{bibliography}

\end{document}